\pdfoutput=1

\documentclass[11pt]{article}

\usepackage[preprint]{acl}

\usepackage{times}
\usepackage{latexsym}

\usepackage[T1]{fontenc}

\usepackage[utf8]{inputenc}

\usepackage{microtype}

\usepackage{inconsolata}

\usepackage{graphicx}
\usepackage{float}          
\usepackage{multirow}   
\usepackage{booktabs}  

\usepackage{stfloats}     

\usepackage{pifont}        

\usepackage{CJKutf8}    

\usepackage[skins]{tcolorbox}  
\tcbuselibrary{breakable}
\usepackage{boites, boites_exemples}

\usepackage{amsmath}

\title{\includegraphics[height=0.75\baselineskip]{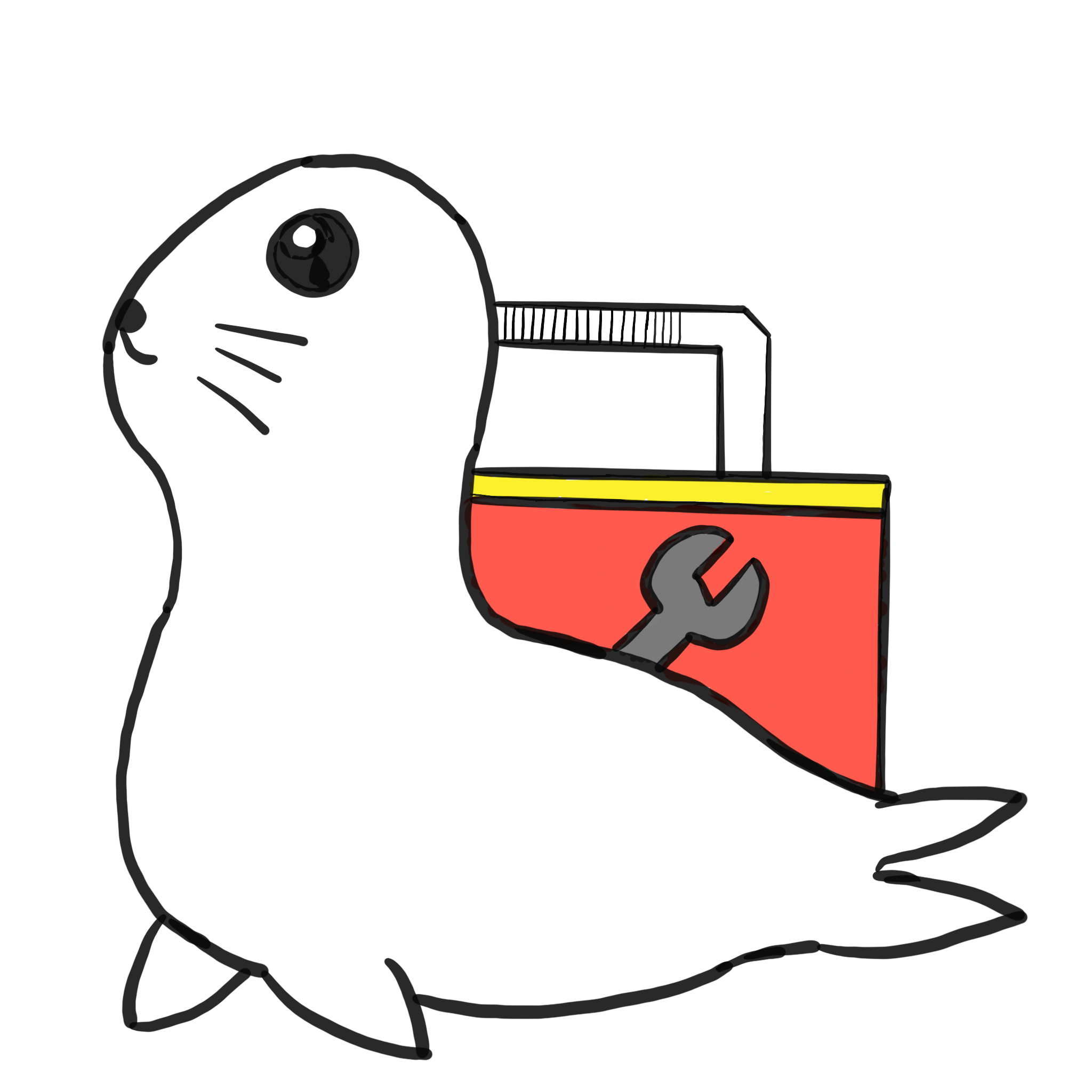} {\color{cyan}Seal-Tools}: Self-Instruct Tool Learning Dataset for Agent Tuning and Detailed Benchmark}

\author{Mengsong Wu \hspace{1cm} Tong Zhu \hspace{1cm} Han Han \hspace{1cm} {\bf Chuanyuan Tan} \\
{\bf Xiang Zhang} \hspace{1cm} {\bf Wenliang Chen} \\
Soochow University, Shizi Street 1, 215006 Suzhou, China \\
  \texttt{\{mswumsw,tzhu7,hhan,cytan17726,xzhangxzhang23\}@stu.suda.edu.cn, wlchen@suda.edu.cn} \\}

\begin{document}
\begin{CJK}{UTF8}{gkai}
\maketitle

\begin{abstract}

This paper presents a new tool learning dataset {\color{cyan}\textbf{Seal-Tools}}, which contains {\color{cyan}\textbf{se}}lf-instruct {\color{cyan}\textbf{A}}PI-{\color{cyan}\textbf{l}}ike tools.
Seal-Tools not only offers a large number of tools, but also includes instances which demonstrate the practical application of tools. 
Seeking to generate data on a large scale while ensuring reliability, we propose a self-instruct method to generate tools and instances, allowing precise control over the process.  
Moreover, our Seal-Tools contains hard instances that call multiple tools to complete the job, among which some are nested tool callings. 
For precise and comprehensive evaluation, we use strict format control and design three metrics from different dimensions.  
Therefore, Seal-Tools can serve as a new benchmark to evaluate the tool-calling ability of LLMs. 
Finally, we evaluate several prevalent LLMs and our finetuned model on Seal-Tools. 
The results show that current systems are far from perfect. 
The code, data and experiment results are available at \url{https://github.com/fairyshine/Seal-Tools} .

\end{abstract}

\section{Introduction}

Large Language Models (LLMs) have shown strong abilities in many tasks in recent years (\citealp{GPT4}, \citealp{ChatGPT_overview}).
Many researchers attempt to use the LLMs as agents (\citealp{HuggingGPT}, \citealp{Gorilla}, \citealp{TaskMatrixAI}), which help users complete difficult tasks by using external tools or plugins. 
The agents serve as a bridge between the users and the tools. 
Therefore, it is particularly important to teach the LLMs how to understand and utilize tools correctly (\citealp{ToolLLM}, \citealp{li-etal-2023-api}). 
For this purpose, we need to prepare high-quality tool learning datasets for enhancing the capabilities of the LLMs as well as precise evaluation.

A tool learning dataset often includes \textbf{tool pool} which contains different kinds of tools, and \textbf{instances} which call the tools to complete tasks. 
In the previous studies, the researchers have built several datasets and achieved a certain of success (\citealp{ToolkenGPT}, \citealp{li-etal-2023-api}, \citealp{ToolBench}), but the datasets exhibit some shortcomings. 
\citet{ToolkenGPT} and \citet{li-etal-2023-api} craft tools by hand but the amount is limited.
\citet{ToolBench} collects real-world APIs from Rapid API Hub\footnote{\url{https://rapidapi.com/hub}} to construct instances but results are evaled by ChatGPT with simple metrics that causes inaccurate evaluation and costs money.
The benchmark of \citet{li-etal-2023-api} is coarse-grained, only considering text similarity during evaluation.
And its training data is not open source.
To summarize, a large-scale high-quality dataset is urgently needed for tuning the agents and performing the automatic precise evaluation.

\begin{figure*}
	\centering   
	\includegraphics[width=0.95\linewidth]{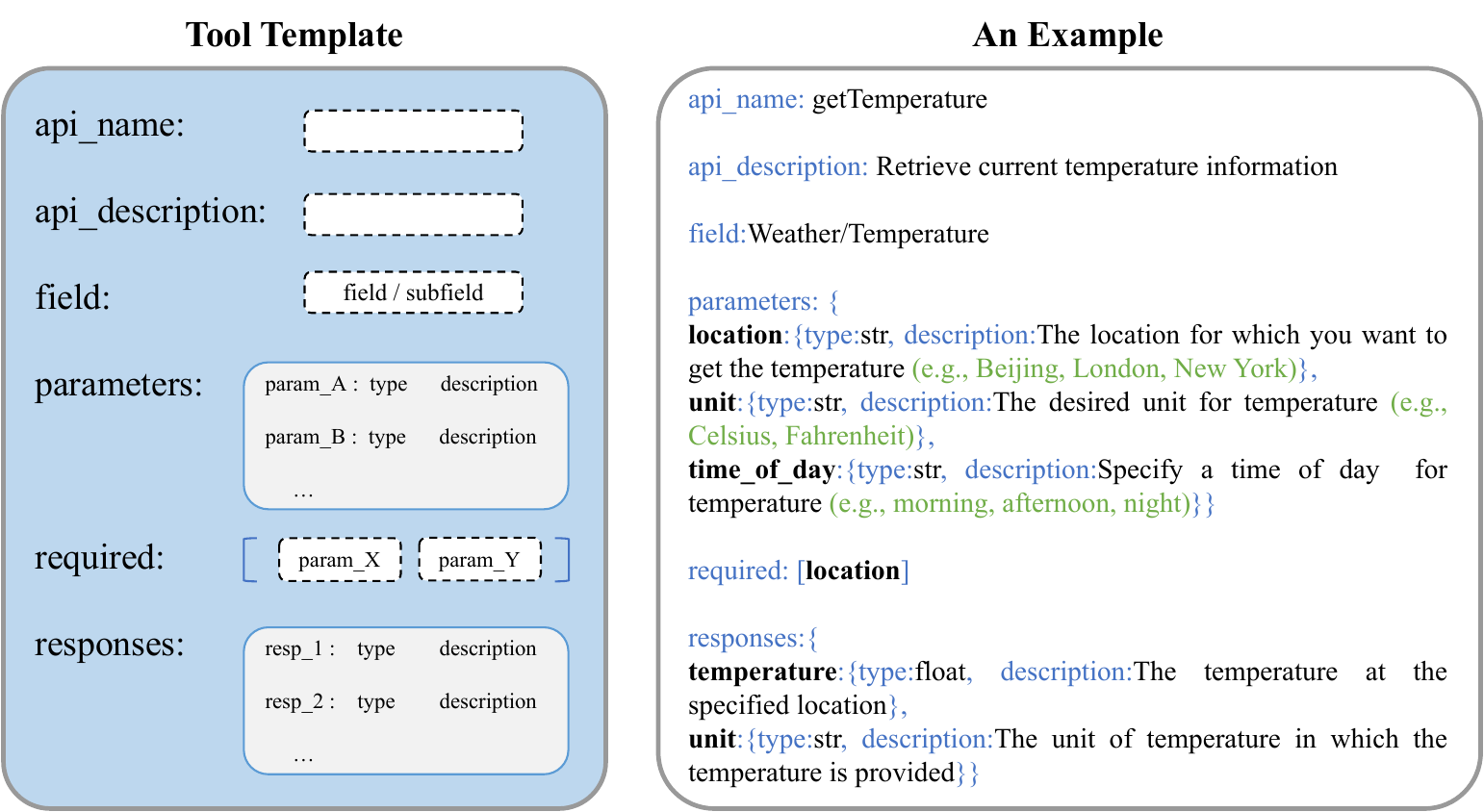}    
	\caption{Tool template of Seal-Tools and the tool "getTemperature" as an example.}    
	\label{tool_template}
\end{figure*}

In our preliminary experiments, we directly use the LLMs to generate the tools and the instances. 
It is easy to obtain a large-scale dataset.
However, the model has limited context length and often outputs duplicate tools. 
The generated instances are mostly simple which can be solved with a quick glance.
Moreover, it’s hard to ensure the correctness of tool callings in each instance due to the LLM hallucination.

To overcome the above challenges, in this paper we propose a self-instruct method to utilize the LLMs to generate a new tool learning dataset, named as \textbf{Seal-Tools}. 
In our method, we first use a LLM to generate a set of fields which refer to different domains, and then tools (just like in Figure~\ref{tool_template}) are generated for each field. 
This simple strategy can well avoid the duplication and diversity problems.
Given the tool pool, we further generate the instances which call single/multiple tools to resolve requests.
We separate the generation into multi steps and set up several checking steps to greatly reduce errors caused by the LLM hallucination.
We use JSON format to describe the tools and instances strictly.
Besides, we successfully generate some instances with nested tool callings thanks to the well-designed construction method and calling template.
These instances have extremely difficult queries to solve and are valuable for finetuning.

To make Seal-Tools a comprehensive benchmark, we design three evaluation dimensions for detailed metrics: Output Format, Tool Selection, and Tool-Parameter Filling-in. 
Since we post-process the LLM outputs into JSON format in Seal-Tools, the evaluation can be more automatic and precise compared to the previous ones.

Contributions of this paper are listed as following:

\begin{itemize}

  \item We propose a self-instruct method to use LLMs to generate tool learning datasets. 
  Our method can generate various in-field tools and single/multiple-tool instances which are mostly reliable through quality control.
  
  \item A brand new tool learning dataset, named as \textbf{Seal-Tools}, is constructed for agent tuning. 
  Compared with the previous datasets, Seal-Tools is relatively large and contains hard instances with nested tool callings. 
  With the help of strict format control, we can perform precise evaluation automatically.

  \item For a comprehensive evaluation, we design 3 main metrics in different dimensions.
  We implement the mainstream agent system and finetune its foundation model with Seal-Tools. 
  From the results, we find that the current systems show room for improvement, especially in nested calling.
\end{itemize}

\section{Related Work}

\subsection{In-Context Learning}

In-context learning (ICL) has become the paradigm for the use of LLMs. 
We add some examples or so-called demonstrations of the task in prompt. 
Models can finish what users want them to do very well through learning from demos. 
According to \citet{a_survey_on_in-context_learning}, ICL presents for the first time in GPT3's technical report (\citealp{GPT3}). 
As it's widely used, \citet{hendel-etal-2023-context} explores the operational mechanisms behind it. 
ICL demos may play similar roles as continuous-value learnable token of prompt tuning. 
\citet{gao2023ambiguity} proves the importance of demo selection and  studied how to select better demos with the help of retriever. 
Auto-ICL (\citealp{Auto-ICL}) attempts to let LLMs generate similar questions with answers. 
ICL then can be applied in scenarios without human supervision.

\subsection{Tool Learning}
\label{sec:toollearning}

Tool learning is actually a subtask of serving LLMs as agents in our opinion.
The agent which supports for tool calling consists of the foundation model, retriever and tool pool. 
The foundation model is the core component which decides how to reply to users and whether to call tools. 
The retriever is responsible for researching relevant tool information according to the user query. 
The tool pool is used to store and manage tool information.
We look through a lot of related works and find that tool learning can be summarized into two categories (prompt-based and finetune-based) according to how LLMs learn to use tools.

\textbf{a. Prompt-Based Tool Learning}

Prompt-based agent has an external tool pool.
The foundation model selects proper tools in prompt given by the retriever like ToolBench (\citealp{ToolLLM}) and  API-Bank (\citealp{li-etal-2023-api}).
HuggingGPT (\citealp{HuggingGPT}) serves models as tools for use.
TaskMatrix.AI (\citealp{TaskMatrixAI}) adds multimodal tools for painting, picture processing and so on.
CREATOR (\citealp{qian-etal-2023-creator}) uses the foundation model to generate tool code  on the spot.
Tool Doc (\citealp{ToolDoc}) replace tool doc with tool demos to improve the performance. 
AgentBench (\citealp{AgentBench}) and TaskWeaver (\citealp{TaskWeaver}) also build up varieties of prompts for tool-calling and do much evaluation in different environments. 
Prompt-based agent has more flexible and large-scale tool pools.
We use this kind of framework in this paper to explore how LLMs can call a huge amount of tools accurately.

\textbf{b. Finetune-Based Tool Learning}

Finetune-based agent doesn't need an extra tool pool. 
The foundation model learns which tools it has and how to use them through finetuning. 
Gorilla (\citealp{Gorilla}) tests the zero-shot scenario (which means no retriever) since it‘s finetuned with relevant self-instruct dataset. 
ToolkenGPT(\citealp{ToolkenGPT}) transforms tools into special tokens and adds them into the vocabulary. 
Toolformer (\citealp{Toolformer}), GeneGPT (\citealp{GeneGPT}) and ToolBench (\citealp{ToolBench}) finetune the foundation model  in a particular scenario. 
ToolBench (\citealp{ToolLLM}) and API-Bank (\citealp{li-etal-2023-api})  build tool learning datasets with plenty of tools. 
Magiccoder~\cite{Magiccoder} finetunes the model with code generation task datasets. 
Finetune-based agent calls tools accurately and rapidly but needs to be finetuned in advance to remember those tools.

\section{Method}

\begin{figure*}[htb]
	\centering   
	\includegraphics[width=0.75\linewidth]{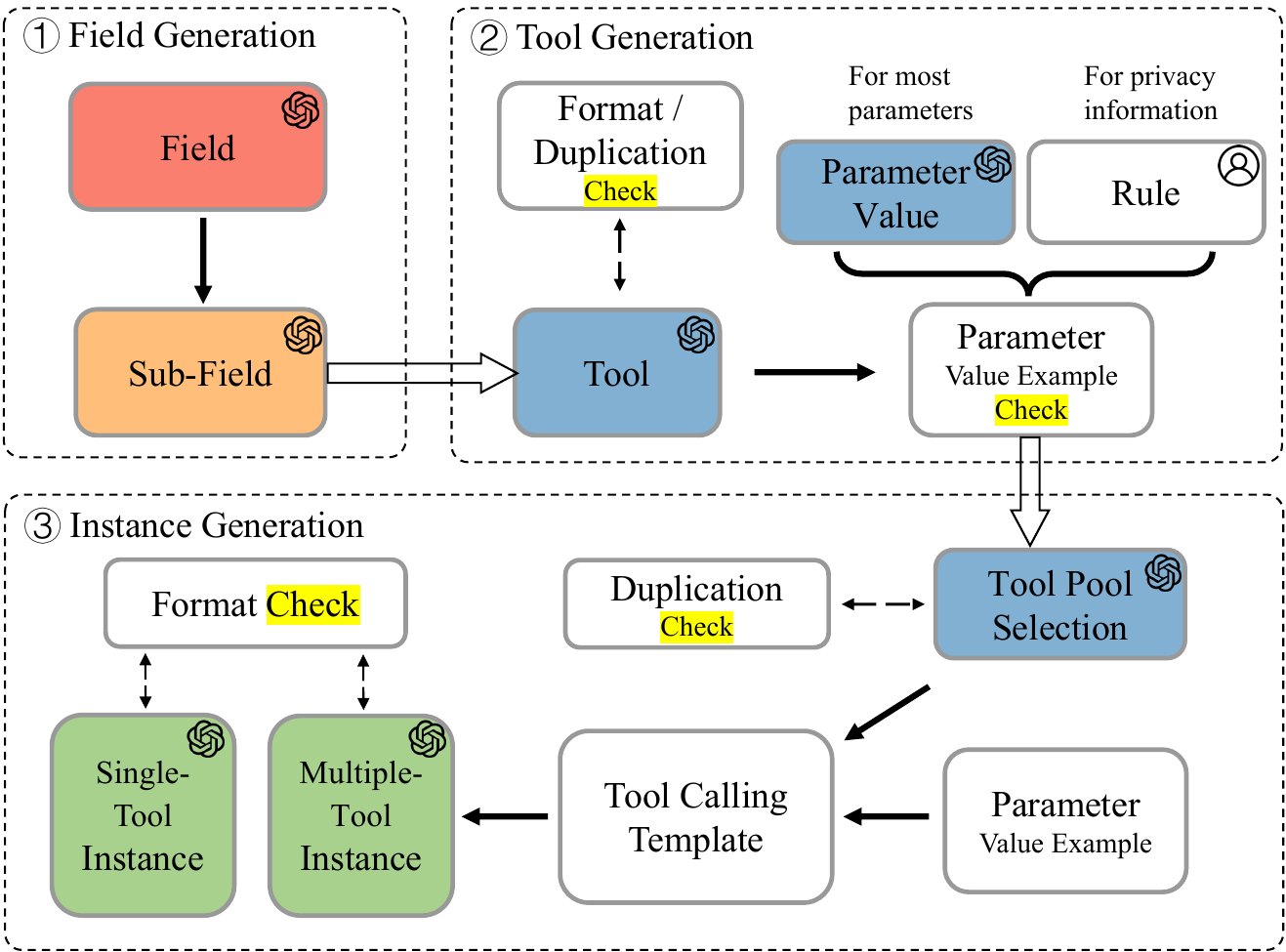}    
	\caption{Flowchart of the dataset construction method.}
	\label{dataset_construction_method}
\end{figure*}

In this section, we talk about how our dataset construction method works and what the quality of constructed result is. 
The highlight of the method is that it is convenient for everyone to try and does not require too much human involvement. 
We can put more efforts on designing the tool template and ICL prompts, LLMs will finish all other jobs.
The amount of data generated can be controlled at will within LLM capability.

\subsection{Dataset Construction}

We adopt the self-instruct strategy to generate datasets with LLMs.
But generative models have shortcomings like hallucination, context length limitations, etc.
When constructing dataset with LLMs, we must make sure there are no logical errors and the answer matches the question because of the hallucination.
Also avoiding too many duplicates when generating data on a large scale is a challenge due to context length limitations.
Thus, to overcome the above challenges, we propose a novel dataset construction method.
As shown in Figure~\ref{dataset_construction_method}, our solution contains three steps:  Field Generation, Tool Generation and Instance Generation.
In our paper, Field, Tool, and Instance are defined as follows:

\textbf{Field}: It describes the specific domain to which the tool belongs. We categorize tools into specific layered fields (2 layers here, field and subfield) based on their use.

\textbf{Tool}: A tool is able to perform a specific job. It has name, description, input/output parameters and so on as shown in Figure~\ref{tool_template} before.

\textbf{Instance}: It is a practical use case of the tool, containing user query and tool calling. There are two categories, \textbf{single-tool instance} and \textbf{multiple-tool instance}. Single-tool instance invokes only one tool while multiple-tool instance invokes multiple tools.

In brief, LLM crafts in-field tools according to pre-generated fields.
Then LLM makes up some instances which can be solved by these tools.

We construct dataset Seal-Tools by ChatGPT.
Due to funding constraints, the dataset scale has not yet reached the upper limit of our construction method and model capacity.

\subsubsection{Field Generation}

This section is about how to generate various fields.
We've tried skipping this step and going straight to tool generation. 
We find the tools generated repeat frequently in that way. 
To enhance tool diversity, we use field information as an anchor. 
Large models are asked to generate tools corresponding to a segmented field. 
We set hierarchical fields to ensure that the functional classification of tools is sufficiently granular.
The number of tools generated has steadily increased. 

\begin{figure}[]    
	\centering   
	\includegraphics[width=0.9\linewidth]{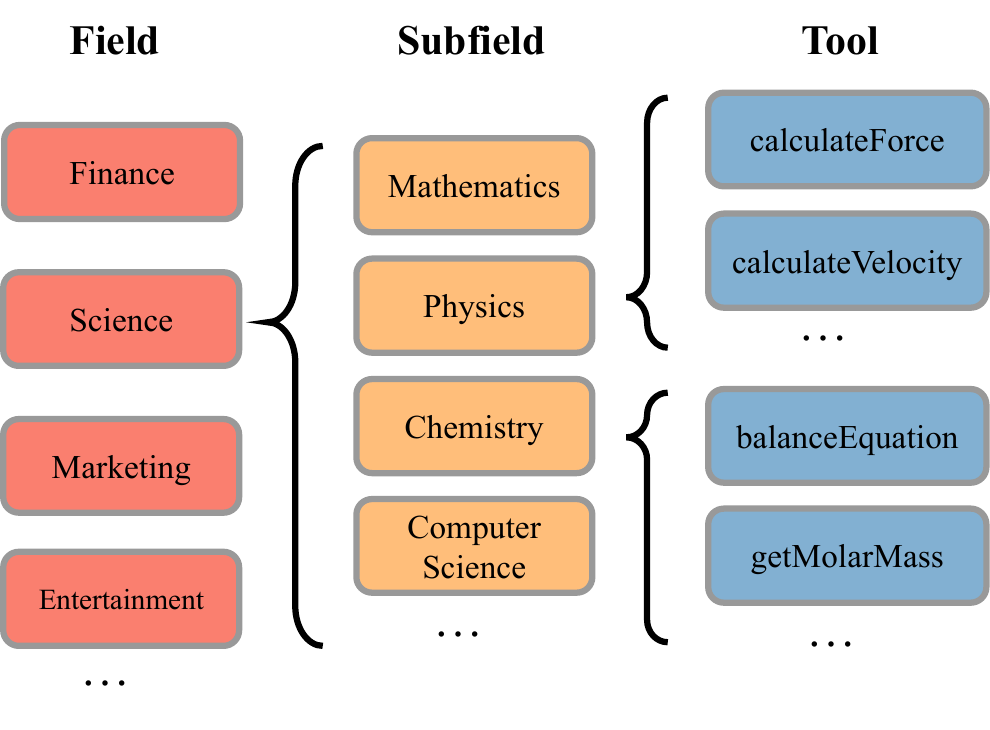}    
	\caption{Some examples of generated fields/subfields and tools.} 
	\label{field_tree}
\end{figure}

First take an field example as the initial demonstration to fill in the prompt. 
LLM generates field set with the insturction "Please generate a field list in the format of a python list." through ICL. 
After that, subfields for all fields in field set are generated with the instruction "Please generate a subfield list in the format of a python list for the \underline{\quad ? \quad} field." in the same way. 
Finally we generate 2 levels of fields, including 146 fields with 5,860 subfields.. 
As shown in Figure~\ref{field_tree}, field ``Science'' has multiple subfields.

\subsubsection{Tool Generation}

Initially, we think this step would be simple and we just have to generate a lot of tools.
In the next step, We try to generate instance with only tool information but it works badly. 
LLM tends to fill in api callings with general and repetitive concepts which are in low quality and these values may be not mentioned in the query.
We later find that it would be effective to generate examples of parameter values at the same time as generating the tool.

When generating parameter values, We find LLM tends to refuse to output entities in real world. 
That's probably a restriction added for security and privacy. 
So we ask LLM to "make up" some proper entities as parameter values. 
The experiment shows that this kind of prompt works well.
But there may be potential privacy leaks from LLMs.

Then we figure out a suitable method for tool generation.
At first, we design the tool template and make up a tool example to initialize the tool pool. 
LLM generates tools in the given subfield with the instruction "Please generate some APIs according to the given field/sub-field. An API is a function with input parameters and output responses."  through ICL. 
When writing the prompt, it's important to add requirements of generating examples of values in parameter description like "(e.g. , \_\_\_, \_\_\_ )". 
After checking the format of generated new tools, we put them into the tool pool if not repetitive.  
When LLM doesn't output new tools in one subfield for several times, we switch to next subfield. 
After generating with all subfields, we examined whether required parameters of each tool have example values. 
If not, we collect them in categories and generate examples in batch through ICL. 
For some parameters of sensitive user information (phone number, email address, etc.) , we generate values using rules. 
In conclusion, we get the tool set and parameter values now.
Finally we generate 4,076 tools. 
Every tool belongs to a subfield just like Figure~\ref{field_tree}.

\begin{figure}[]    
	\centering   
	\includegraphics[width=0.8\linewidth]{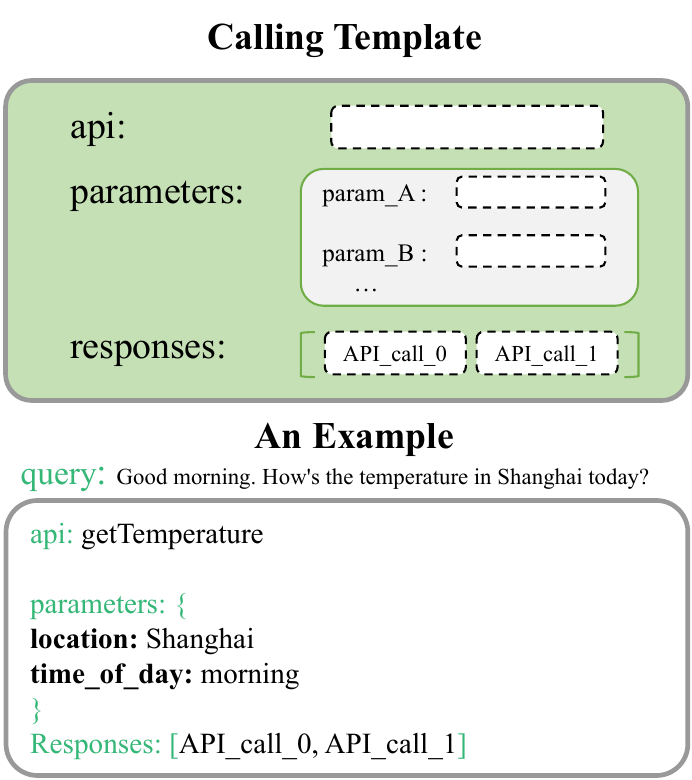}    
	\caption{An instance template for single-tool calling.} 
	\label{calling_template}
\end{figure}

\subsubsection{Instance Generation}

The instance contains two parts: the user query and tool callings (how to invoke tools like Figure~\ref{calling_template}) . 
LLMs can make up different styles of queries with given parameter values. 
For better finetuning effects, we choose to generate queries in a brief style. 

For single-tool instances, we choose the tool and pick up parameter values to fill in the prompt. 
It is successful to generate the single-tool invoking instance by ICL. 
Finally we generate 4076 single-tool instances.

For multiple-tool instances, the prompt in single-tool instance generation is hard to get good output.
The tool callings generated through ICL are mostly confusing in format.
Since it might be difficult for LLMs nowadays, we try to simplify it into two steps and transform the question answering task into blank filling task. 
At the first step, LLM should choose several tools which can be combined into one query from huge amount candidate tools in prompt as shown in Figure~\ref{dataset_construction_method}. 
Then we generate the tool invoking template according to the chosen tools. 
At the second step, LLM only needs to fill in parameter values with given examples and generate the relevant query.
Finally we generate 10k multiple-tool instances.

\begin{figure}[htb]
	\centering   
	\includegraphics[width=0.99\linewidth]{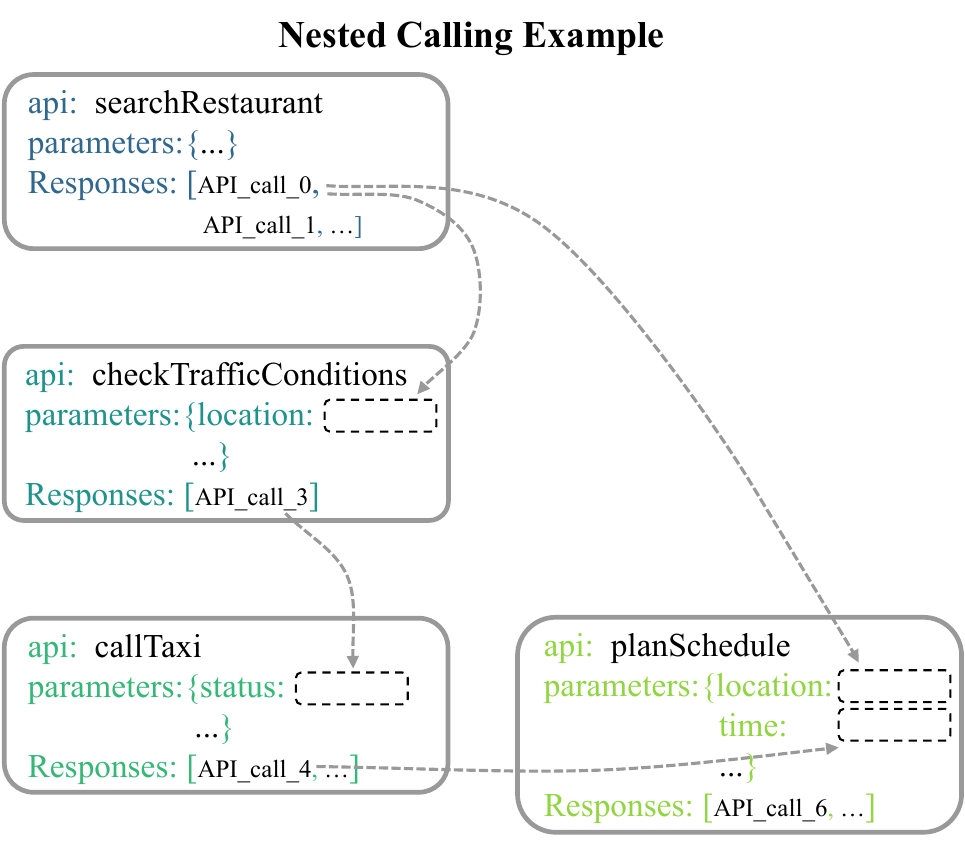}    
	\caption{An example for multiple-tool instance with nested calling.}
	\label{nested_calling}
\end{figure}

The innovation of this step is that we could generate some multiple-tool instances with nested calling (simply called \textbf{nested instances}) in this way. 
In the nested instance, Figure~\ref{nested_calling} for example, the response value of the previous invoked tool could be parameter value of the next invoked tool. 
All tools invoked can form a directed acyclic graph instead of flow line. 
This is closer to complex real-world application scenarios and makes our evaluations more effective.

\begin{table*}[htb]
\centering
\renewcommand{\arraystretch}{1.0}
\resizebox{0.98\textwidth}{!}{
\begin{tabular}{clcccc}
\toprule
\multicolumn{2}{c}{\textbf{}}                                  & \multicolumn{1}{c}{\begin{tabular}[c]{@{}c@{}}\textbf{Seal-Tools}\\(ours)\end{tabular}} & \multicolumn{1}{c}{\begin{tabular}[c]{@{}c@{}}\textbf{ToolBench}\\(\citealp{ToolLLM})\end{tabular}} & \multicolumn{1}{c}{\begin{tabular}[c]{@{}c@{}}\textbf{API-Bank}\\(\citealp{li-etal-2023-api})\end{tabular}} & \multicolumn{1}{c}{\begin{tabular}[c]{@{}c@{}}\textbf{ToolAlpaca}\\(\citealp{ToolAlpaca})\end{tabular}} \\ \midrule
\multirow{3}{*}{\textbf{Tools}}     & \textbf{Source}                 & self-instruct                           & real-world                                       & self-instruct + annotation            & real-world + annotation                 \\
                                    & \textbf{Amount}                 & 4,076                                    & 16,464                                            & 2,211                                  & 2,386                                    \\
                                    & \textbf{Avg. params (required)} & 1.551                                   & 1.013                                            & unknown                               & N/A$^{\dagger}$                \\ \midrule
\multirow{5}{*}{\textbf{Instances}} & \textbf{Source}                 & self-instruct                           & self-instruct                                    & self-instruct + annotation            & self-instruct                           \\
                                    & \textbf{Amount}                 & 14,076                                   & 126,486                                           & 4,125                                  & 3,938                                    \\
                                    & ├\textbf{Multiple-tool callings}          & 10,000                                   & $\approx$85,330                                   & 615                                   & 1,426                                    \\
                                    & └\textbf{Nested-tool callings}         & 586                                     & N/A$^{\clubsuit}$                                   & unknown                               & N/A$^{\clubsuit}$                          \\ 
                                    & {\textbf{Cross-field tool callings}}        & {\color[HTML]{008114}\ding{52}}   &  {\color[HTML]{008114}\ding{52}}           & unknown                               &  {\color{red}\ding{56}}                                     \\\midrule
\multirow{3}{*}{\textbf{Benchmark}}  & \textbf{Metric}                 & Acc, P/R/F1                          & Pass Rate, Win Rate                          & Acc, Rouge           & N/A                           \\
									& \textbf{Tool parameter}                 & {\color[HTML]{008114}\ding{52}}                           & {\color{red}\ding{56}}                                    & {\color[HTML]{008114}\ding{52}}            & N/A                           \\
                                    & \textbf{Deterministic evaluation}      & {\color[HTML]{008114}\ding{52}}       & {\color{red}\ding{56}}                                      & {\color[HTML]{008114}\ding{52}}            & N/A                          \\ \midrule
\multicolumn{2}{l}{\textbf{Total open-source?}}                       & {\color[HTML]{008114}\ding{52}}                                     &  {\color[HTML]{008114}\ding{52}}                                             & {\color{red}\ding{56}}                               &  {\color[HTML]{008114}\ding{52}}                                     \\
\multicolumn{2}{l}{\textbf{Extensible (fully self-instruct) ?}}          &  {\color[HTML]{008114}\ding{52}}                                     &{\color{red}\ding{56}}                                               & {\color{red}\ding{56}}                                   & {\color{red}\ding{56}}                                     \\ \bottomrule
\end{tabular}
}
\caption{Comparison of several Tool Learning datasets.  $^{\dagger}$ Formatting confusion. $^{\clubsuit}$ Multi-step.}
\label{dataset_comparsion}
\end{table*}

\subsection{Dataset Analysis}

We compare Seal-Tools with several common tool learning datasets in Table~\ref{dataset_comparsion}: 
ToolBench\footnote{\textbf{ToolBench} lists data in Table 1 of the original paper. 
We count number of APIs  as tools in our settings.}, 
API-Bank\footnote{\textbf{API-Bank} has evaluation data implemented by human and self-instruct training data. 
We get information of APIs and single/multiple callings in Table 2 of the original paper. 
The training data seems to be not publicly available.} 
and ToolAlpaca\footnote{\textbf{ToolAlpaca} lists data in Table 1 of the original paper. 
We count number of functions as tools.}.
The result shows that our dataset is competitive among all datasets.

\begin{figure}[htb]    
	\centering   
	\includegraphics[width=0.99\linewidth]{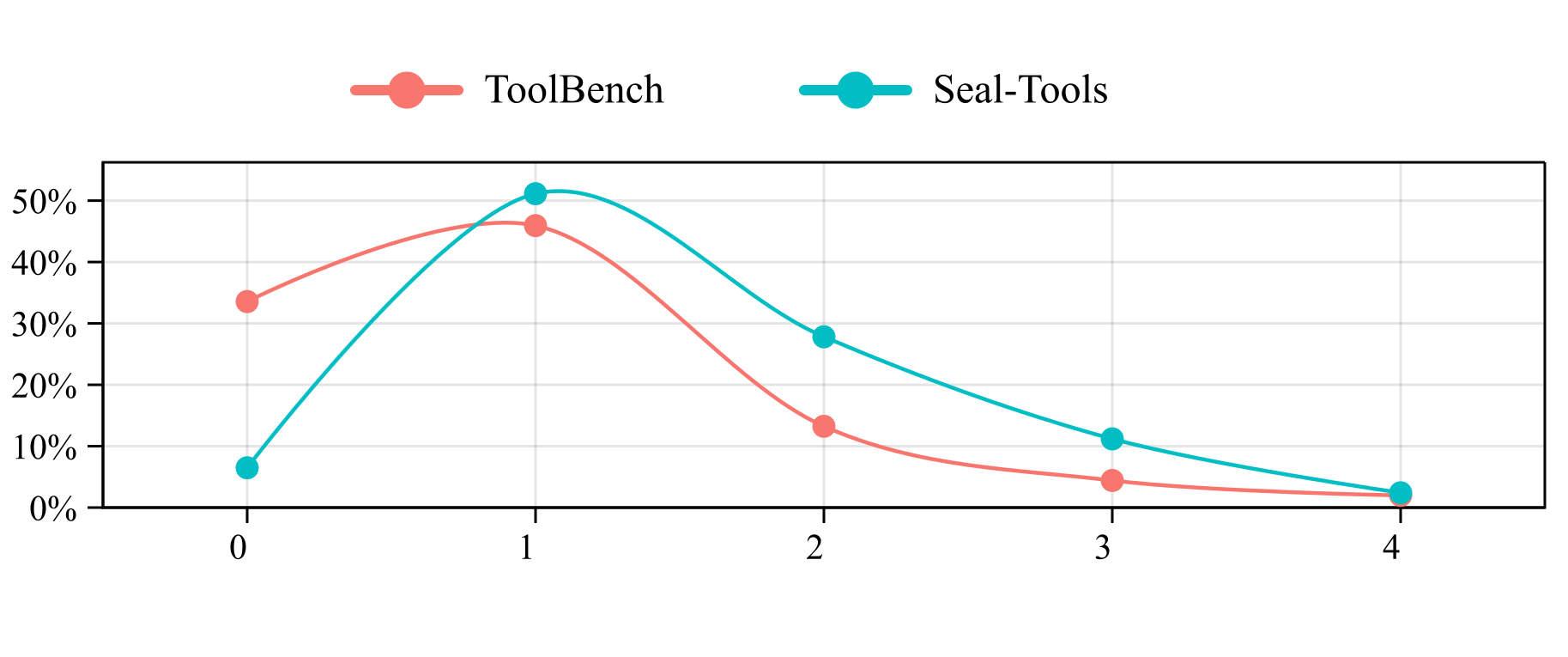}    
	\caption{Required parameters amount of tools comparison between \textbf{Seal-Tools} and \textbf{ToolBench}.}
	\label{average_param}
\end{figure}

With regard to tools, we propose the first open-source batch generation method with self-instruct strategy.
The quality of tools generated may be not inferior to real-world tools collected by ToolBench.
As Figure~\ref{average_param} shows, we find nearly 34\% tools in ToolBench have no required parameters. In Seal-Tools, the amount is only 6\%. 
Models have to fill in more parameters in each instances just like the avg. params amount show in Table~\ref{dataset_comparsion}.

As for instances, the scale of Seal-Tools is very large, second only to ToolBench.
It also contains more hard instances like cross-field callings and nested callings which test LLM's ability to think logically.
Most datasets ask LLMs to call one tool in one response.
It often needs multiple steps to handle difficult problems.
While Seal-Tools asks LLMs to give above multiple-tool callings in one turn chat instead of step-by-step by other datasets.
It is much more difficult to solve and closer to real-world scenarios.
We think it helps improve agent execution efficiency.

Seal-Tools also provides very detailed evaluations.
We post-process the output of LLMs into JSON format for evaluating tool selection and parameter filling-in.
But ToolBench and API-Bank do not process it.
ToolBench could only get pass rate and win rate given by ChatGPT.
It is not deterministic since ChatGPT has its own preferences and limitations.
API-Bank calculates the rouge score.
ToolAlpaca doesn‘t provide evaluation.
When evaluating agents, we could calculate the correctness of every selected tool and its parameters.
Model capabilities such as understanding tools, generating parameters, etc. are presented more clearly.

Moreover, it can be extended with our current most automated construction method.
Seal-Tools can be used in more places without worrying about dataset scale.
We just need to provide the universal tool template and ICL prompts to generate much more data.
Other method can only generate more instances.
Both tools and instances can be generated much more within the capability of LLMs.
Furthermore, tools and instances are generated together in a complete self-consistent generative process in our method so they can be more harmonized with each other.

\section{Experiment}

\subsection{Evaluation Metric}

There are currently no standardized evaluation metrics for tool learning task.
For evaluating in detail, we design 3 main eval metrics: Format ACC, Tool P/R/F1 and Parameter P/R/F1. 
Specific calculations can be found in Appendix \ref{appx:metric}.

\textbf{Format ACC}  measures the format accuracy of model output. 
The foundation model should follow instructions in the prompt so that tools can be invoked well. 

\textbf{Tool P/R/F1} measures the tool selection ability of foundation model.
The specific calculation is similar to P/R/F1 metrics for the information extraction task.
Please refer to the appendix for details, the same below.

\textbf{Parameter P/R/F1} measures the tool parameter filling-in ability.  
We use these metrics to hierarchically examine the capability of the foundation model to understand and use tools.

\subsection{Main Result}
\label{sec:main result}

\subsubsection{Overall}

\begin{table*}[]
\centering
\resizebox{0.88\textwidth}{!}{
\begin{tabular}{lccccccc}
\toprule
\multicolumn{1}{c}{\multirow{2}{*}{\textbf{Model}}} & \multicolumn{1}{c}{\multirow{2}{*}{\textbf{Format ACC}}} & \multicolumn{3}{c}{\textbf{Tool}}                                                                 & \multicolumn{3}{c}{\textbf{Parameter}}                                                            \\
\multicolumn{1}{c}{}                                & \multicolumn{1}{c}{}                                     & \multicolumn{1}{c}{\textbf{P}} & \multicolumn{1}{c}{\textbf{R}} & \multicolumn{1}{c}{\textbf{F1}} & \multicolumn{1}{c}{\textbf{P}} & \multicolumn{1}{c}{\textbf{R}} & \multicolumn{1}{c}{\textbf{F1}} \\
\cmidrule(lr){1-1} \cmidrule(lr){2-2} \cmidrule(lr){3-5} \cmidrule(lr){6-8} 
\textbf{ChatGPT} \emph{(gpt-3.5-turbo-0613)}                                             & 96.16                                                     & 83.20                          & 74.73                          & 78.74                            & 68.63                          & 66.85                          & 67.73                           \\
\textbf{GPT4} \emph{(gpt-4-0613)}                                                & 97.12                                                     & 90.02                          & 74.71                          & 81.65                            & 80.52                          & 67.57                          & 73.48                      \\ 
\midrule
\textbf{LLaMA2} \emph{7B}                                           & 40.55                                                     & 47.91                          & 26.74                          & 34.33                            & 33.52                          & 20.43                          & 25.39                           \\
\textbf{LLaMA2-Chat} \emph{7B}                                       & 78.73                                                     & 62.10                          & 53.91                          & 57.72                            & 44.92                          & 43.24                          & 44.06                           \\
\textbf{Vicuna} \emph{7B-v1.5}                                              & 70.83                                                     & 67.33                          & 49.81                          & 57.26                            & 49.11                          & 42.26                          & 45.43                           \\
\textbf{Mistral} \emph{7B-Instruct-v0.2}                                             & 77.03                                                     & 76.84                          & 59.65                          & 67.16                            & 64.81                          & 50.25                          & 56.61                           \\
\textbf{ToolLLaMA2} \emph{7B-v2}                                            & 13.44                                                     & 19.35                          & 0.96                           & 1.84                             & 18.98                          & 0.84                           & 1.61                            \\
\textbf{Ours} \emph{(finetuned on LLaMA2-7B)}                                 &                                                           &                                &                                &                                  &                                &                                &                                 \\
\hspace{1em}w/ BM25                                               & 95.57                                                     & 79.67                          & 74.79                          & 77.15                            & 73.51                          & \textbf{70.76}                          & 72.11                           \\
\hspace{1em}w/ DPR                                                & \textbf{95.86}                                                     & \textbf{82.81}                          & \textbf{77.84}                          & \textbf{80.25}                            & \textbf{75.95}                          & 70.23                          & \textbf{72.98}                           \\
\bottomrule   
\end{tabular}
}
\caption{Overall result. 
All LLMs use DPR retriever as default.
}
\label{result_agent}
\end{table*}

Here is the main evaluation section of the paper. 
We hope that this experiment simulates the performance of agent in a real environment. 
As mentioned in Section \ref{sec:toollearning}, when a prompt-based agent system receives a user query, it uses the retriever to search for relevant tools firstly. 
Then the foundation model decides whether to use candidate tools according to the query and generate a reply. 
So we eval the performance of different retrievers and select the best retriever DPR to add to the system. 
More details about the selection of retriever are in Appendix~\ref{appx:retriever}.

We finetune LLaMA2-7B\footnote{\url{https://huggingface.co/meta-llama/Llama-2-7b-hf}} with Seal-Tools.
The tools in prompt are given mainly by retriever DPR.
We add the missing gold tools to the prompt in the train split.
When evaluating, DPR gives the top-5 relevant tools.

Table~\ref{result_agent} represents the main result of tool learning task. 
Open-source LLMs perform similarly. 
They all have great potential for making further progress.
Searchers can try to add more tool learning datasets in the pre-training phase.

It is also reasonable that ToolLLaMA which is finetuned with ToolBench has a bad performance.
Single dataset fine-tuning severely impacts LLM's understanding of other instructions.

After finetuning with Seal-Tools, our model can output much more correct tool callings than the base model.
Tool F1 increases 45.92\% and Parameter F1 increases 47.59\% as shown in Table~\ref{result_agent}.
It even outperforms ChatGPT and is slightly inferior to GPT4. 
The score proves that Seal-Tools is efficient in finetuning.

\begin{figure}[htb]
	\centering   
	\includegraphics[width=0.98\linewidth]{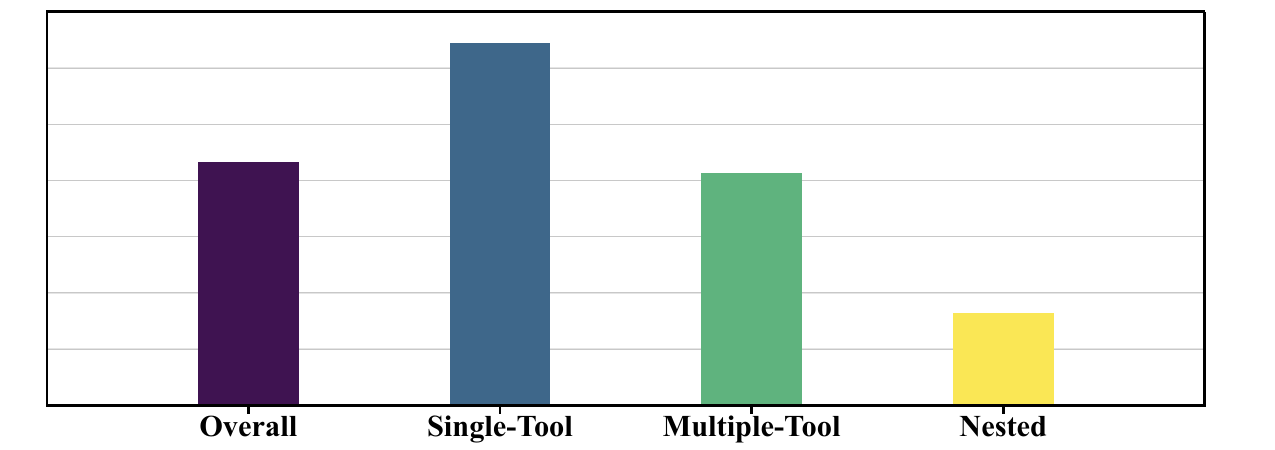} 
	\caption{Simplicity level of different kinds of instances.
	Visualized using Mistral's Parameter F1 metric.}
	\label{result_difficulty}
\end{figure}

We discuss about various instances in Section \ref{sec:single/multiple} and Section \ref{sec:nested} below.
The difficulty of different instances is shown in Figure~\ref{result_difficulty} roughly for reference (Higher in the bar graph means simpler).

\subsubsection{Single/Multiple-Tool Instances}
\label{sec:single/multiple}

\begin{table*}[]
\centering
\resizebox{0.98\textwidth}{!}{
\begin{tabular}{lcccccccccccccc}
\toprule
\multicolumn{1}{c}{\multirow{3}{*}{\textbf{Model}}} & \multicolumn{7}{c}{\textbf{Single-Tool}}                                                                                                                                                    & \multicolumn{7}{c}{\textbf{Multi-Tool}}                                                                                                                                  \\
\cmidrule(lr){2-8} \cmidrule(lr){9-15} 
\multicolumn{1}{c}{}                       & \multicolumn{1}{c}{\multirow{2}{*}{\textbf{Format ACC}}} & \multicolumn{3}{c}{\textbf{Tool}}                         & \multicolumn{3}{c}{\textbf{Parameter}}                    & \multicolumn{1}{c}{\multirow{2}{*}{\textbf{Format Acc}}} & \multicolumn{3}{c}{\textbf{Tool}}                         & \multicolumn{3}{c}{\textbf{Parameter}} \\
\multicolumn{1}{c}{}                       & \multicolumn{1}{c}{}                                     & \textbf{P} & \textbf{R} & \multicolumn{1}{c}{\textbf{F1}} & \textbf{P} & \textbf{R} & \multicolumn{1}{c}{\textbf{F1}} & \multicolumn{1}{c}{}                                     & \textbf{P} & \textbf{R} & \multicolumn{1}{c}{\textbf{F1}} & \textbf{P}  & \textbf{R} & \textbf{F1} \\
\cmidrule(lr){1-1} \cmidrule(lr){2-2} \cmidrule(lr){3-5} \cmidrule(lr){6-8} \cmidrule(lr){9-9} \cmidrule(lr){10-12} \cmidrule(lr){13-15}
\textbf{ChatGPT}                                     & 98.98                                                     & 88.01      & 94.90      & 91.33                            & 74.28      & 83.94      & 78.82                            & 95.38                                                     & 82.70      & 73.01      & 77.55                            & 68.11       & 65.49      & 66.77       \\
\textbf{GPT4}                                        & 98.64                                                     & 88.16      & 96.26      & 92.03                            & 82.00      & 85.16      & 83.55                            & 96.70                                                     & 90.24      & 72.86      & 80.62                            & 80.37       & 66.17      & 72.58      \\ 
\midrule
\textbf{LLaMA2}                                   & 44.22                                                     & 25.83      & 42.18      & 32.04                            & 15.93      & 28.66      & 20.48                            & 39.53                                                     & 54.52      & 25.42      & 34.68                            & 38.43       & 19.78      & 26.11       \\
\textbf{LLaMA2-Chat}                              & 85.37                                                     & 40.27      & 81.63      & 53.93                            & 26.54      & 63.21      & 37.38                            & 76.89                                                     & 67.02      & 51.54      & 58.27                            & 49.03       & 41.64      & 45.03       \\
\textbf{Vicuna}                                      & 76.53                                                     & 47.65      & 72.45      & 57.49                            & 33.79      & 59.76      & 43.17                            & 69.25                                                     & 71.13      & 47.88      & 57.23                            & 51.85       & 40.87      & 45.71       \\
\textbf{Mistral}                                     & 86.73                                                     & 72.99      & 86.39      & 79.13                            & 66.14      & 68.29      & 67.20                            & 74.34                                                     & 77.36      & 57.36      & 65.88                            & 64.67       & 48.81      & 55.63       \\
\textbf{ToolLLaMA}                                   & 21.77                                                     & 12.50      & 2.72       & 4.47                             & 11.94      & 1.63       & 2.86                             & 11.13                                                     & 22.95      & 0.81       & 1.57                             & 21.05       & 0.78       & 1.50        \\
\textbf{Ours}                                        &                                                           &            &            &                                  &            &            &                                  &                                                           &            &            &                                  &             &            &             \\
\hspace{1em}w/ BM25                                      & \textbf{98.30}                                                     & 91.81      & 91.50      & 91.65                            & 84.31      & 85.16      & 84.73                            & 94.81                                                     & 78.57      & 73.36      & 75.87                            & 72.61       & \textbf{69.61}      & 71.08       \\
\hspace{1em}w/ DPR                                       & \textbf{98.30}                                                     & \textbf{93.13}      & \textbf{92.18}      & \textbf{92.65}                            & \textbf{85.54}      & \textbf{85.37}      & \textbf{85.45}                            & \textbf{95.19}                                                     & \textbf{81.88}      & \textbf{76.61}      & \textbf{79.16}                            & \textbf{75.12}       & 69.02      & \textbf{71.94}       \\
\bottomrule
\end{tabular}
}
\caption{Results for single-tool / multiple-tool instances.
}
\label{result_1/n}
\end{table*}

We count for single/multiple-tool instances in Table~\ref{result_1/n} for more details.
For most models, they do better in single-tool instances than multi-tool instances.
Calling a single tool is easier than calling multiple tools, which is to be expected.
While LLaMA2 and Vicuna are exceptions.
Since this looks a little strange, we check their outputs and find that they tend to call more tools when dealing with single-tool instances.
Maybe LLaMA2 uses some low-quality corpus during pre-training or it is not well trained. 
LLaMA2 tends to invoke more tools and doesn't know how to make trade-offs.

Our model outperforms GPT4 in single-tool instances but falls slightly behind in multiple-tool instances.
How to further improve the performance of multiple-tool calling is the focus of our future research.

\subsubsection{Nested Instances}
\label{sec:nested}

\begin{table}[htb]
\centering
\resizebox{0.48\textwidth}{!}{
\begin{tabular}{lccccccc}
\toprule
\multicolumn{1}{c}{\multirow{2}{*}{\textbf{Model}}} & \multicolumn{1}{c}{\multirow{2}{*}{\textbf{Format ACC}}} & \multicolumn{3}{c}{\textbf{Tool}}                         & \multicolumn{3}{c}{\textbf{Parameter}} \\
\multicolumn{1}{c}{}                                & \multicolumn{1}{c}{}                                     & \textbf{P} & \textbf{R} & \multicolumn{1}{c}{\textbf{F1}} & \textbf{P}  & \textbf{R} & \textbf{F1} \\
\cmidrule(lr){1-1} \cmidrule(lr){2-2} \cmidrule(lr){3-5} \cmidrule(lr){6-8}
LLaMA2-Chat                                       & 79.86                                                     & 73.04      & 58.39      & 64.90                            & 37.23       & 34.66      & 35.90       \\
Mistral                                       & 68.43                                                     & 84.16      & 57.67      & 68.44                            & 52.00       & 36.94      & 43.20       \\
\textbf{Ours}                                                 & 96.76                                                     & 89.64      & 85.82      & 87.69                            & 77.32       & 74.15      & 75.70       \\
├ has seen (501)                                 & 96.41                                                     & 91.03      & 86.61      & 88.76                            & 78.88       & 75.43      & 77.12       \\
└ still unseen (85)                              & 98.82                                                     & 81.71      & 81.08      & 81.40                            & 67.66       & 66.02      & 66.83      \\
\bottomrule
\end{tabular}
}
\caption{Result for nested instances.
}
\label{result_nested}
\end{table}

We collect all 586 nested instances in Seal-Tools.
The performance of different models is listed in Table~\ref{result_nested}.
It shows that nested instances are the most difficult for models to solve.
LLaMA2-Chat and Mistral behaves poorly compared to other data.

For our model, since 501 of these data have been used in the train split, We calculate the final scores separately.
Although the model has seen them during the finetuning process, its Parameter F1 is only 77.12\%, not too high.
For unseen data, it performs 5.11\% worse than multiple-tool instances.
But it is still better than raw model LLaMA2 which confirms the effect for finetuning.
In summary, our dataset Seal-Tools is of high quality with these hard instances.

\subsection{Extended Result}

\subsubsection{Evaluation of Parameter Filling-In}
\label{sec:evaluation on parameter}

We focus on testing LLMs' parameter filling-in ability for multiple-tool instances in this sub-section.
Prompts for finetuning LLaMA2 and evalutation only contain gold tools.
The results are listed in Table~\ref{result_parameter_filling-in}.

For tool selection, since only gold tools are given in prompt, Tool P for each model should ideally reach 100\%.
Our model works as expected, but LLaMA2 and ChatGPT do not due to the in-context hallucination. 
For parameter filling in, as shown in Table~\ref{result_parameter_filling-in}, the finetuned model performs very well even through only the results of multiple-tool instances are counted.
Considering all the above facts,, Seal-tools is helpful for improving the ability of parameter filling-in for the LLM.
However, how to finetune the LLM to make it more robust remains to be investigated.

\begin{table}[htb]
\centering
\resizebox{0.48\textwidth}{!}{
\begin{tabular}{lccccccc}
\toprule
\multicolumn{1}{c}{\multirow{2}{*}{\textbf{Model}}} & \multicolumn{1}{c}{\multirow{2}{*}{\textbf{Format ACC}}} & \multicolumn{3}{c}{\textbf{Tool}}                         & \multicolumn{3}{c}{\textbf{Parameter}} \\
\multicolumn{1}{c}{}                                & \multicolumn{1}{c}{}                                     & \textbf{P} & \textbf{R} & \multicolumn{1}{c}{\textbf{F1}} & \textbf{P}  & \textbf{R} & \textbf{F1} \\
\cmidrule(lr){1-1} \cmidrule(lr){2-2} \cmidrule(lr){3-5} \cmidrule(lr){6-8}
LLaMA2-chat                                       & 82.74                                                     & 99.86      & 80.72      & 89.27                            & 72.37       & 70.38      & 71.36       \\
ChatGPT                                              & 94.06                                                     & 99.97      & 92.82      & 96.26                            & 80.94       & 85.22      & 83.02       \\
\textbf{Ours}                                                 & \textbf{98.87}                                                     & \textbf{100.00}     & \textbf{98.84}      & \textbf{99.41}                            & \textbf{94.26}       & \textbf{93.65}      & \textbf{93.95}    \\
\bottomrule  
\end{tabular}
}
\caption{Result of parameter filling-in. Only needed tools are given in the prompt.}
\label{result_parameter_filling-in}
\end{table}

\subsubsection{Error Analysis}

\begin{figure}[htb]
	\centering   
	\includegraphics[width=0.98\linewidth]{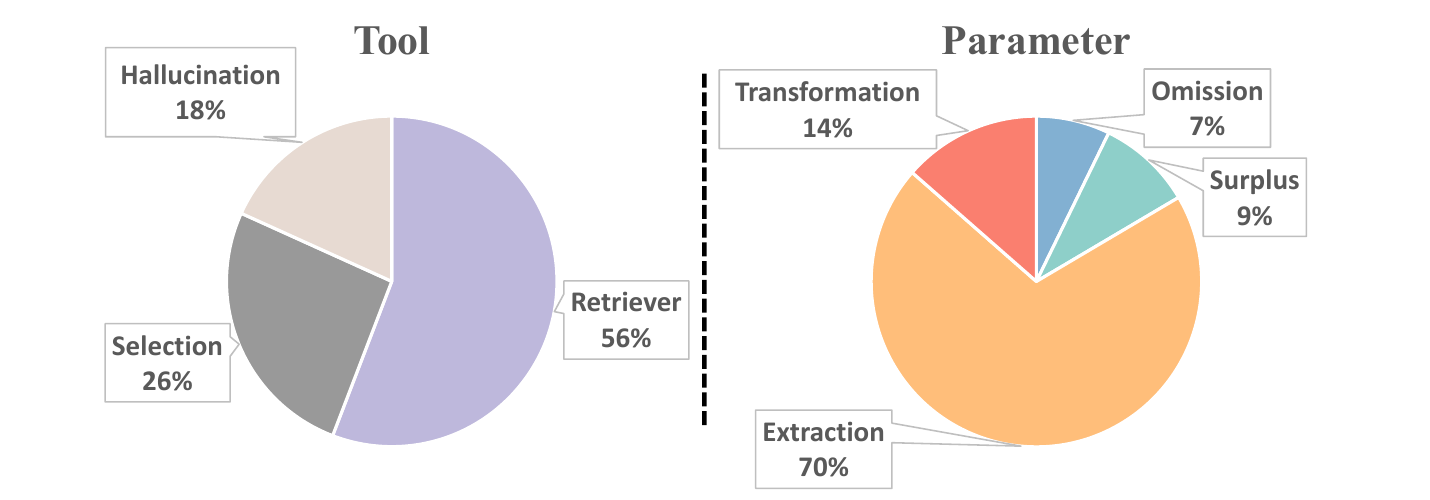} 
	\caption{Error in tool selection and parameter filling-in.}
	\label{error_analysis}
\end{figure}

In Figure~\ref{error_analysis}, we count the types of errors made by our model with retriever DPR in Section \ref{sec:main result}.
Understanding where mistakes are made allows us to continually strive for further improvement.

For tool selection, how to retrieve all needed tools is most urgent. 
The limitations of existing retriever training methods are mentioned in Appendix~\ref{appx:retriever}.
Hallucination also needs to be noted, since models may generate tools that are not in the retrieval results.

For parameter filling in, most of the errors are that models do not extract the correct keywords from queries.
LLM omits the required parameters for 7\% and overfills with unmentioned parameters for 9\%.
Besides, 14\% of the errors are due to models not understanding the query requirements or not converting to the parameter request format.

\section{Conclusion}

In this paper, we present a novel construction method for building dataset Seal-Tools which includes a set of tools and instances. 
In our method, we carefully control the quality of auto-generated data, increasing reliability and diversity. 
In Seal-Tools, hard instances include nested tool callings and cross-field callings, which have seldomly been investigated in previous studies.. 
We further design evaluation metrics from three dimensions. 
Experimental results show that current agent systems still have room for improvement. 
We believe that Seal-Tools can serve as a new benchmark and boost the research on tool learning with LLMs.

\bibliography{custom}

\appendix

\section{Prompts of Dataset Construction}

\subsection{Field Generation}

\begin{boiteepaisseavecuntitre}{Generating Initial Field}
\noindent Please generate a field list in the format of a python list. Try to cover all areas.

\noindent Tips:

\noindent 1. The field should be coarse-grained.

\noindent 2. The job is really important. Please finish it perfectly with your full effort.

\noindent For example: 

\noindent field\_list = [

\noindent	"\{\}",
	
\noindent	"\{\}",
	
\noindent]
\end{boiteepaisseavecuntitre}

\begin{boiteepaisseavecuntitre}{Generating Sub-Field}
\noindent Please generate a subfield list in the format of a python list for the "\{\}" field.

\noindent Tips:

\noindent 1. The subfield should be fine-grained.

\noindent 2. The subfield list is used to classify tasks to the specified subfield.

\noindent 3. The job is really important. Please finish it perfectly with your full effort.
\end{boiteepaisseavecuntitre}

\subsection{Tool Generation}

\begin{boiteepaisseavecuntitre}{Generating In-Field Tool}
\noindent Please generate some APIs according to the given field/sub-field. An API is a function with input parameters and output responses. It's like a tool to help with all kinds of fields. The generated APIs should be related to the field. This task is really important for human beings, so please finish it with your best effort.

~\\

\noindent For example:

\noindent field:"\{\}"

\noindent sub-field:"\{\}"

\noindent \{\}

~\\

\noindent Tips:

\noindent 1. Generate enough parameters in "parameters" list. Parameters in the "required" list are definitely needed each time; only core parameters should be selected from "parameters" list.

\noindent 2. The format of the "field" key is "field/sub-field".

\noindent 3. Your answer should be in JSON format，the format of your answer should be strictly consistent with the example.

\noindent 4. Make sure descriptions of parameters end with examples of values in the format of "(e.g., value1, value2, value3, ...)". You can just make up some.

\noindent 5. The "type" key in lists of "parameters" and "responses" should be selected from ["str", "int", "float", "bool"].

~\\

\noindent Now generate some APIs like above as many as possible.

~\\

\noindent field:"\{\}"

\noindent sub-field:"\{\}"

\{\{ \}\}
\end{boiteepaisseavecuntitre}

\subsection{Instance Generation}

\begin{boiteepaisseavecuntitre}{Generating Single-Tool Instance}
\noindent Please generate the task description. The task requires calling API to finish. Make sure the task description coherent and natural. Please don't mention API in task description, API calling should be obtained by logical derivation.

~\\

\noindent For example:

\noindent function calling =

\noindent \{\{"api":"translate", "parameters":\{\{"text":"Hello world", "source\_language":"English", "target\_language":"Japanese"\}\}\}\}

\noindent Task description =

\noindent [Tell me how to speak "Hello world" in Japanese.]

~\\

\noindent function calling =

\noindent \{\{"api":"book\_meeting", "parameters":\{\{"meeting\_title":"academic research", "meeting\_date":"2023-09-10", "meeting\_time":"3:00 p.m."\}\}\}\}

\noindent Task description =

\noindent [Book a meeting for "academic research" on September 10, 2023, at 3:00 p.m.]

~\\

\noindent Now finish the following content in the format of the example above. 

~\\

\noindent function calling =

\noindent \{\}

\noindent Task description =

\noindent [ ]
\end{boiteepaisseavecuntitre}

\begin{boiteepaisseavecuntitre}{Tool combination}
\noindent Here is a list of APIs. Please select and combine parts of given\_apis to create a specific and complex task. 

\noindent Tips:

\noindent 1. Ensure that the selected APIs have a strong association and a logical relationship to each other. 

\noindent 2. You don't need to follow the original order of the APIs, but the chronological order of execution.

~\\

\noindent For example:

\noindent input:

\noindent given\_apis = [\{\{'getWeatherForecast': 'Retrieve weather forecast information'\}\}, \{\{'calculateBMI': 'Calculate Body Mass Index (BMI) based on height and weight'\}\}, \{\{'translateText': 'Translate text from one language to another'\}\}, \{\{'generateQRCode': 'Generate a QR code for a given text or URL'\}\}, \{\{'getHotelDetails': 'Retrieve detailed information about a hotel'\}\}, \{\{'getAirQualityIndex': 'Retrieve the air quality index (AQI) information for a specific location'\}\}, \{\{'searchRestaurant': 'Search for a restaurant based on various criteria'\}\}, \{\{'checkTrafficConditions': 'Retrieve current traffic conditions information'\}\}, \{\{'searchHotels': 'Search for hotels based on various criteria'\}\}, \{\{'reserveRentalCar': 'Reserve a rental car for a specific location and time'\}\}, \{\{'checkFlightAvailability': 'Check the availability of flights for a specified route and date'\}\}, \{\{'getArticleDetails': 'Retrieve details of an article by providing its identifier'\}\}, \{\{'cancelHotelReservation': 'Cancel a hotel reservation'\}\}, \{\{'callTaxi': 'Request a taxi service for transportation'\}\}]

\noindent output:

\noindent selected\_apis = ['searchHotels', 'getHotelDetails', 'cancelHotelReservation']

\noindent task\_description = ['Find the reserved hotel and obtain its information in order to cancel the reservation due to a schedule change.']

~\\

\noindent Please return the chosen list of APIs in the format of a Python list named 'selected\_apis' and generate a paragraph describing the task, as shown in the upper example. Don't mention any API in the 'task\_description'.

~\\

\noindent input:

\noindent given\_apis = \{\}

\noindent output:
\end{boiteepaisseavecuntitre}

\begin{boiteepaisseavecuntitre}{Generating Multiple-Tool Instance}
\noindent Please use APIs in api\_list to create a specific and complex task. First, fill in the blanks with parameter values in api\_calling. Then, write the task description based on the api calling.

\noindent Tips for improved\_api\_calling generation:

\noindent 1. Borrow from the parameter description or make up some specific and niche entities in reality as parameter values, without using the word "example". 

\noindent 2. Whenever possible, use the responses("API\_call\_" + serial number) of previous APIs as parameter values.

\noindent 3. For different parameters, try to set the same value to combine APIs together and make the task more consistent.

\noindent Tips for task\_description generation:

\noindent 1. Make sure that all parameter values in the improved\_api\_calling list are mentioned in the task\_description except the "API\_call\_" + serial number or "API".

~\\

\noindent For example:

\noindent input:

\noindent api\_list = [\{\{"api\_name": "searchRestaurant", "api\_description": "Search for a restaurant based on various criteria", "parameters": \{\{"cuisine": \{\{"type": "str", "description": "The type of cuisine you prefer"\}\}, "price\_range": \{\{"type": "str", "description": "The price range you're looking for"\}\}, "rating": \{\{"type": "float", "description": "The minimum rating you want for the restaurant"\}\}, "open\_now": \{\{"type": "bool", "description": "Specify if you want to find restaurants that are currently open (true or false)"\}\}\}\}, "required": [], "responses": \{\{"location": \{\{"type": "str", "description": "The location of the enquired restaurant"\}\}\}\}\}\}, \{\{"api\_name": "checkTrafficConditions", "api\_description": "Retrieve current traffic conditions information", "parameters": \{\{"location": \{\{"type": "str", "description": "The location for which you want to check traffic conditions"\}\}, "time\_of\_day": \{\{"type": "str", "description": "Specify the time of day for checking traffic conditions"\}\}, "traffic\_source": \{\{"type": "str", "description": "Specify the source of traffic information"\}\}, "include\_incidents": \{\{"type": "bool", "description": "Include information about traffic incidents and accidents"\}\}\}\}, "required": ["location"], "responses": \{\{"traffic\_level": \{\{"type": "str", "description": "The traffic level at the specified location"\}\}, "estimated\_travel\_time": \{\{"type": "int", "description": "The estimated travel time in minutes based on current traffic conditions"\}\}, "average\_speed": \{\{"type": "int", "description": "The average speed of traffic in miles per hour (mph)"\}\}, "incidents": \{\{"type": "str", "description": "Information about any traffic incidents or accidents (if included in the request)"\}\}\}\}\}\}, \{\{"api\_name": "callTaxi", "api\_description": "Request a taxi service for transportation", "parameters": \{\{"pickup\_location": \{\{"type": "str", "description": "The location where you want to be picked up"\}\}, "destination": \{\{"type": "str", "description": "The destination address where you want to go"\}\}, "passenger\_count": \{\{"type": "int", "description": "The number of passengers"\}\}, "ride\_type": \{\{"type": "str", "description": "The type of ride you prefer"\}\}, "special\_requests": \{\{"type": "str", "description": "Any special requests or instructions for the driver"\}\}\}\}, "required": ["pickup\_location", "destination"], "responses": \{\{"status": \{\{"type": "str", "description": "The status of the taxi request"\}\}, "driver\_name": \{\{"type": "str", "description": "The name of the assigned taxi driver (if available)"\}\}, "estimated\_arrival\_time": \{\{"type": "str", "description": "The estimated time of arrival of the taxi"\}\}\}\}\}\}]

\noindent origin\_api\_calling = [\{\{"api": "searchRestaurant", "parameters": \{\{"cuisine": \_\_\_\}\}, "responses": ["API\_call\_0"]\}\}, \{\{"api": "checkTrafficConditions", "parameters": \{\{"location": \_\_\_, "time\_of\_day": \_\_\_\}\}, "responses": ["API\_call\_1", "API\_call\_2", "API\_call\_3", "API\_call\_4"]\}\}, \{\{"api": "callTaxi", "parameters": \{\{"pickup\_location": \_\_\_, "destination": \_\_\_\}\}, "responses": ["API\_call\_5", "API\_call\_6", "API\_call\_7"]\}\}]

\noindent output:

\noindent improved\_api\_calling = [\{\{"api": "searchRestaurant", "parameters": \{\{"cuisine": "Italian"\}\}, "responses": ["API\_call\_0"]\}\}, \{\{"api": "checkTrafficConditions", "parameters": \{\{"location": "API\_call\_0", "time\_of\_day": "afternoon"\}\}, "responses": ["API\_call\_1", "API\_call\_2", "API\_call\_3", "API\_call\_4"]\}\}, \{\{"api": "callTaxi", "parameters": \{\{"pickup\_location": "Nanjing Road", "destination": "API\_call\_0"\}\}, "responses": ["API\_call\_5", "API\_call\_6", "API\_call\_7"]\}\}]

\noindent task\_description = ["Please help me to plan a convenient and enjoyable dinner outing. Find a nearby Italian restaurant with good reviews, then check the current traffic conditions from Nanjing Road to the restaurant"s location. If the traffic is favorable, you can reserve a rental car at 5:00 p.m. for the evening."]

~\\

\noindent Please complete the following content in the provided format above. You only need to return the "improved\_api\_calling" list and the "task\_description".

\noindent input:

\noindent api\_list = \{\}

\noindent origin\_api\_calling = \{\}

\noindent output:
\end{boiteepaisseavecuntitre}

\section{Generated Tool Amount per Time}
\label{appx:generatiton_amount}

\begin{figure}[htb]
	\centering   
	\includegraphics[width=0.9\linewidth]{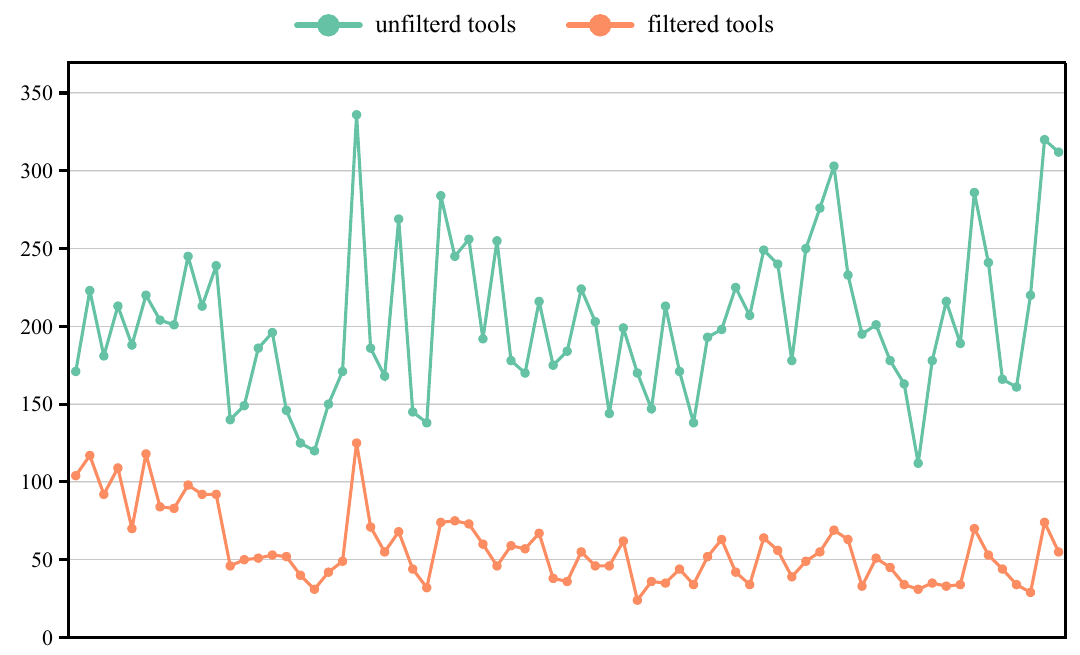} 
	\caption{Generated Tool amount per 200 times. Duplicate tools are filtered in real-time.}    
	\label{tool_generation}
\end{figure}

In the whole process, The amount of generated tools holds up like in Figure~\ref{tool_generation}.
Benefits from the pre-generated field information, tools can be generated on a larger scale.
Otherwise we may get only around 100 tools by ChatGPT.

\section{Detailed Formulae for Evaluation Metrics}
\label{appx:metric}

\begin{align*}
    &\resizebox{\linewidth}{!}{%
    $\begin{aligned}
    Format \thinspace ACC & = \frac{amount_{correct \thinspace format}}{amount_{all}} \\
    Tool \thinspace P & = \frac{amount_{correct \thinspace tools}}{amount_{predict \thinspace tools}} \\
    Tool \thinspace R & = \frac{amount_{correct \thinspace tools}}{amount_{gold \thinspace tools}} \\
    Tool \thinspace F1 & = \frac{2 \cdot Tool \thinspace P \cdot Tool \thinspace R}{Tool \thinspace P + Tool \thinspace R} \\
    Parameter \thinspace P & = \frac{amount_{correct \thinspace tools}}{amount_{predict \thinspace tools}} \\
    Parameter \thinspace R & = \frac{amount_{correct \thinspace tools}}{amount_{gold \thinspace tools}} \\
    Parameter \thinspace F1 & = \frac{2 \cdot Parameter \thinspace P \cdot Parameter \thinspace R}{Parameter \thinspace P + Parameter \thinspace R}
    \end{aligned}$}
\end{align*}

\section{Retriever Comparison}
\label{appx:retriever}

Two classical retrievers are tested, the discrete retriever BM25 and the dense retriever DPR. 
There is something special when training DPR. 
Retrievers like DPR are used in open-domain QA task before. 
Generally speaking, a open-domain question may have multiple answers but it's okay to answer only one of them. 
However in the tool learning task, the retriever is asked to find out all needed tools. 
Even one missing tools means that the reply can't be generated properly. 
The loss function in training step is different from that in traditional settings. 
We try to use contrastive loss but the result is terrible. 
The gold tools of one query is trained one by one and the retriever seems to only remember the last tool seen. 
We use the ranking loss finally and get significantly improved result. 
There should be better prescription remained to be studied. 

\begin{figure}[htb]
	\centering   
	\includegraphics[width=0.9\linewidth]{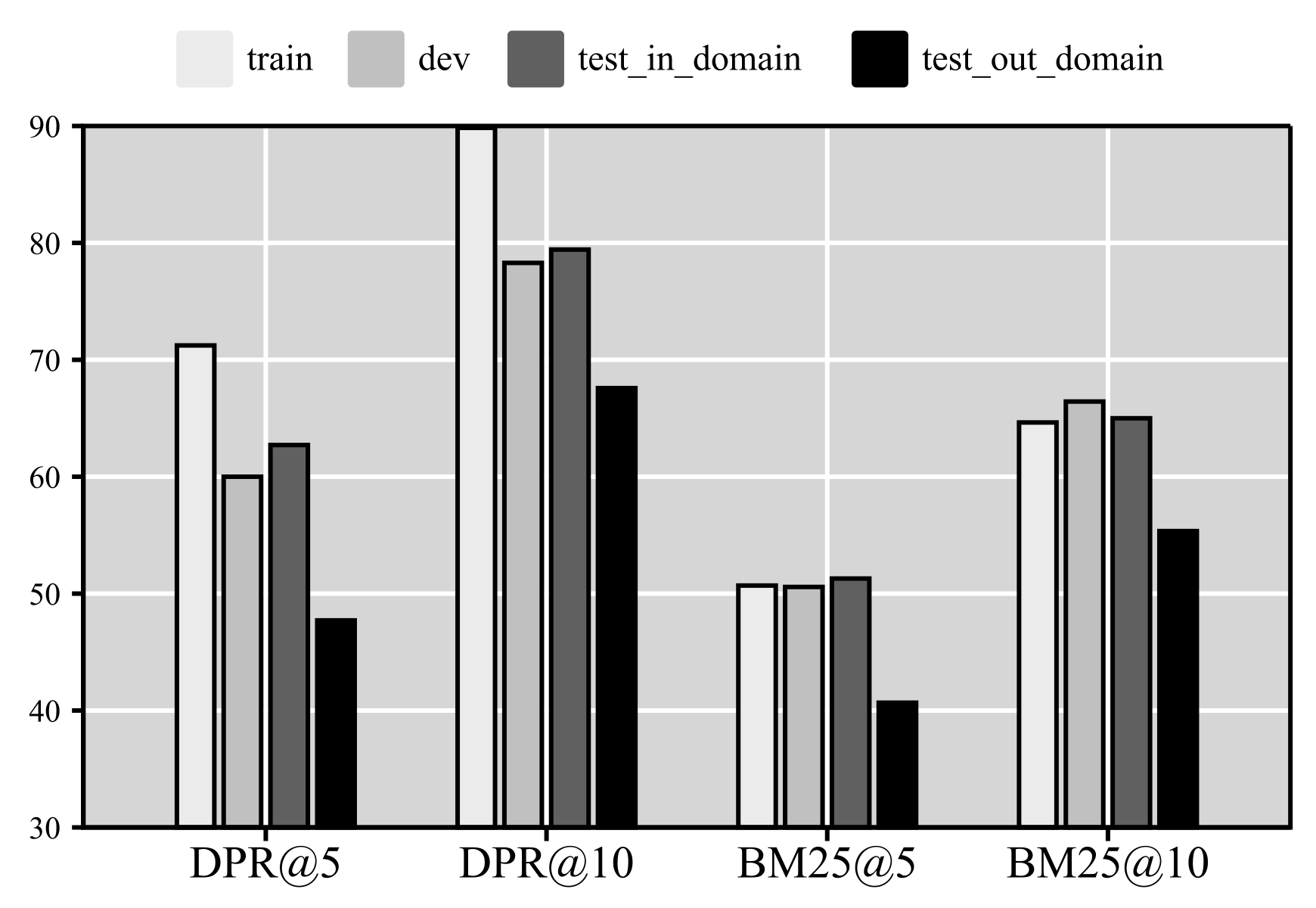} 
	\caption{Retriever: DPR v.s. BM25.}    
	\label{result_retriever}
\end{figure}

As the result in Figure~\ref{result_retriever} shows, DPR performs better than BM25 which means the retrieval task involves semantic understanding. 
We can't find out all relevant tools by simple keywords matching. 
We use DPR as the retriever in agent system finally. 
How to optimize the retriever deserves further research.

\section{Generalization Verification}
\label{appx:4 prompts}

\begin{table}[htb]
\centering
\resizebox{0.49\textwidth}{!}{
\begin{tabular}{llll}
\toprule
Model           & \multicolumn{1}{c}{Tool\_ACC}               & \multicolumn{1}{c}{Param\_Selection\_F1}    & \multicolumn{1}{c}{Param\_Fill-in\_F1}      \\  \midrule
LLaMA2-Chat     & \textbf{55.49}/\textbf{26.78}/13.49/25.05 & \textbf{48.80}/35.55/17.76/29.83 & 27.79/24.02/16.51/23.61 \\
Ours & 32.56/26.58/\textbf{27.75}/\textbf{31.60} & 42.87/\textbf{40.20}/\textbf{42.85}/\textbf{44.02} & \textbf{32.30}/\textbf{31.76}/\textbf{35.93}/\textbf{33.01} \\  \bottomrule
\end{tabular}
}
\caption{Experiments on Benchmark which is transformed from TOD datasets.}
\label{result_TOD}
\end{table}

We construct a small tool learning benchmark for initial validation before. 
In the early stages of exploring the tool learning task, we think it's a good idea to get useful data from other tasks since there is a serious lack of  it. 
This dataset is transformed from task-oriented dialogues (TOD) datasets SGD \citep{SGD}, Multwoz2.2 \citep{zang-etal-2020-multiwoz} and Taskmaster \citep{byrne-etal-2019-taskmaster}. 
The dataset contains 20 tools and 519 instances. 

We design 4 styles of prompts (vanilla, role-play, task description and imitaton) to eval through ICL in Table~\ref{result_TOD}:

\textbf{Vanilla}: It's the basic instruction containing a brief introduction to the task.

\textbf{Role-play}: We let LLM play the role of an NLP technologist who is well versed in all kinds of tasks.
The Prompt focuses on describing the powerful natural language understanding capabilities of the technologist and does not describe the task specifics.

\textbf{Task description}: In such prompts, we inform LLMs in detail how to fulfill the corresponding tasks.
We explain the key concepts involved in detail and describe how to perform them step by step.

\textbf{Imitation}: Given LLM's strong in-context learning or few-shot learning capabilities, this prompt focuses on guiding LLM to learn how to complete the task from the examples given.

It's somewhat unexpected that the model finetuned in Section \ref{sec:evaluation on parameter} performs better than the chat model. 
Our Seal-Tools might be helpful in instruction tuning to enhance the overall capabilities of LLMs. 
But when we test the finetuned model in Section \ref{sec:main result}, it perfroms badly.
We might get out some conclusions: the model learns how to fill in parameters when given all needed tools in prompt; the model learns how to select tools and what tools are used for when given the retrieved tool list.

\end{CJK}
\end{document}